# WEB IMAGE ANNOTATION BY DIFFUSION MAPS MANIFOLD LEARNING ALGORITHM

Neda Pourali[1]

[1]Department of Electronic, Computer and IT, Science and Research Branch, Islamic Azad University, Qazvin, Iran

## ABSTRACT

*Automatic image annotation is one of the most challenging problems in machine vision areas. The goal of this task is to predict number of keywords automatically for images captured in real data. Many methods are based on visual features in order to calculate similarities between image samples. But the computation cost of these approaches is very high. These methods require many training samples to be stored in memory. To lessen this burden, a number of techniques have been developed to reduce the number of features in a dataset. Manifold learning is a popular approach to nonlinear dimensionality reduction. In this paper, we investigate Diffusion maps manifold learning method for web image auto-annotation task. Diffusion maps manifold learning method is used to reduce the dimension of some visual features. Extensive experiments and analysis on NUS-WIDE-LITE web image dataset with different visual features show how this manifold learning dimensionality reduction method can be applied effectively to image annotation.*

## 1. INTRODUCTION

Duo to the explosive growth of digital technologies, many visual data are created and stored. There is an urgent need of effective and efficient method to detect visual information of data. In many areas of machine learning, pattern recognition, information retrieval and machine vision, we are dealing with high-dimensional data. To search and classify such huge image collections poses significant technical challenges. Many approaches have been proposed to solve the image annotation problem, including classification-based [1, 2], region-based [3, 4, 5], graph-based [6, 7], topic model [8, 9], and generative image patch modelling [10] techniques. Some previous studies have shown that a simple non-parametric nearest neighbor-like approach can be successful. For example, Makadia [11] proposed JEC method which exploits multiple visual features to improve accuracy of task. The main issue for success of these methods is to employ rich visual features with appropriate distance metrics defined in raw feature spaces. So image presentation can become high-dimensional. For example Guillaumin [12] worked on 37,000 dimensional data.

While these approaches are successful from the view of annotation accuracy, their computational costs, in terms of both memory use and complexity. Moreover, to realize a practical large-scale system with acceptable response speed, implementation of an approximate nearest neighbour search method will be necessary. In some studies, to lessen this burden, a number of techniques have been developed to vastly reduce the complexity of these non-parametric techniques. The original Euclidian distance as the similarity measure in the input feature space can be approximated by the Hamming distance. The training costs of these methods, however, are basically expensive. Another popular problem with high dimensional data is known as the "curse of dimensionality". So we need some methods to reduce the dimension of feature vectors.

Manifold learning is an emerging and promising approach in non-parametric dimension reduction. A manifold is a topological space that is locally Euclidean, i.e. around every point, there is a neighborhood that is topologically the same as the open unit ball in $R^n$. An Earth is a good example of a manifold. Locally, at each point on the surface of the Earth, we have a 3-D coordinate system: two for location and one for the altitude. Globally, it is a 2-D sphere that is embedded in a 3-D space [13].

Manifolds offer a powerful framework for dimensionality reduction problem. The key idea of dimensionality reduction is to find low dimensional structure that is embedded in a higher dimensional space. Occam's razor has been used to support dimensionality reduction. The main idea of Occam's razor is to select the simplest model from a set of equal models to explain a given phenomenon. Moreover, if the data are indeed generated according to a manifold, then a manifold based learning can be optimal [13].

In recent studies, Manifold learning algorithms have been used for many applications. Nathan Mekuz [14] worked on Local Linear Embedding (LLE) method for face recognition problem. Raducanu [15] represented Supervised Laplacian Eigenmaps (S-LEM) in order to enhance accuracy of classification task on some face datasets. Carlotta Orsenigo [16] proposed manifold learning methods for cancer microarray data classification. Abdenour Hadid [17] worked on demographic classification from face videos using manifold learning. Yahong Han [18] proposed manifold learning application for image classification on out of sample data. Xianming Liu [19] worked on cross-media manifold learning for image retrieval and annotation task.

In this paper, we investigate an advanced manifold learning method for image auto-annotation task. Local Tangent Space Alignment is used to reduce the dimension of some visual features. After reducing the dimension of feature vectors, new features are used for our multi-label classifier. We compare these methods with some other dimensionality reduction methods.

The rest of this paper is organized as follows: Section 2 introduces dimensionality reduction and Section 3 explains Diffusion maps manifold learning method and Diffusion maps algorithm is used to solve the curse of dimensionality problem of KNN classifier. Section 4 introduces experimental setup and experimental results. At the end of this paper, we make a conclusion in Section 5.

## 2. DIMENSIONALITY REDUCTION

The problem of nonlinear dimensionality reduction method can be defined as follows. Assume we have dataset representation in a $n \times D$ matrix $X$ consisting of n datavectors $x_i (i \in \{1, 2, \ldots, n\})$ with dimensionality $D$. Assume that this dataset has intrinsic dimensionality of d (where $d < D$, and often we have $d \ll D$). Intrinsic dimensionality means that the points in dataset $X$ are lying on or near a manifold with dimensionality of d that is embedded in the $D$-dimensional space. Dimensionality reduction problems map the dataset $X$ with dimensionality $D$ into a dataset $Y$ with dimensionality $d$ while retaining the geometry of the data as much as possible. Neither the geometry of the data manifold, nor the intrinsic dimensionality $d$ of the dataset $X$ are known. So, dimensionality reduction is a problem that can only be solved by assuming certain properties of the data. Throughout the paper, datapoint is denoted by $x_i$, where $x_i$ is the $i$'th row of the $D$- dimensional data matrix $X$. The low-dimensional of $x_i$ is denoted by $y_i$, where $y_i$ is the $i$'th row of the $d$-dimensional data matrix $Y$. [20].

Figure (1) shows taxonomy of techniques for dimensionality reduction. The main distinction between techniques for dimensionality reduction is the distinction between linear and nonlinear techniques. Linear techniques assume that the data lie on or near a linear sub-space of the high-dimensional space. Nonlinear techniques for dimensionality reduction do not rely on the linearity assumption as a result of which more complex embedding of the data in the high-dimensional space can be identified. In section 3 we discuss Diffusion maps nonlinear dimensionality reduction.

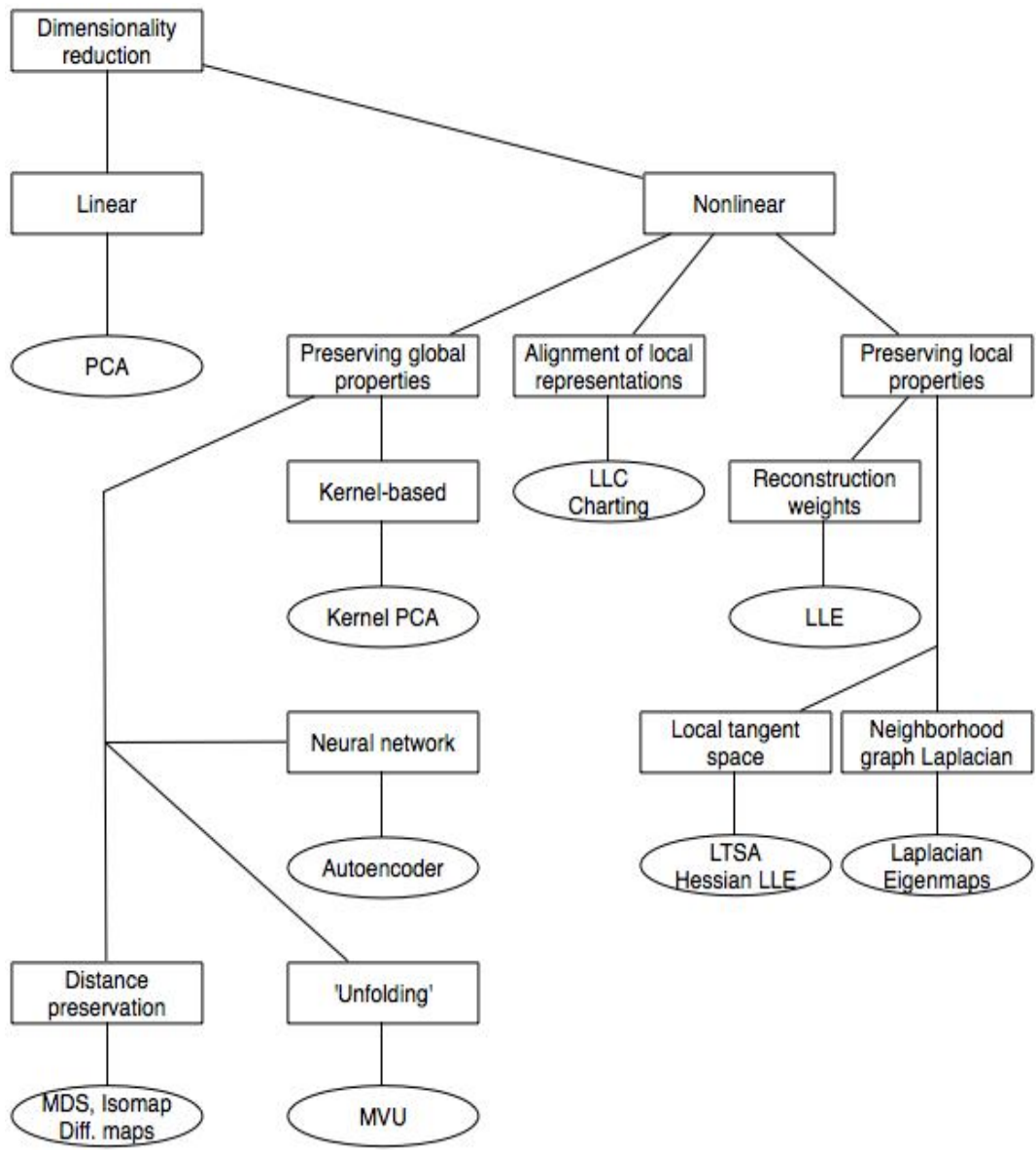

Figure 1.Taxonomy of dimensionality reduction techniques [20].

## 3. DIFFUSION MAPS MANIFOLD LEARNING DIMENSIONALITY REDUCTION

The diffusion maps (DM) algorithm [21, 22] originates from the dynamical systems. Maaten [20] introduced LEM as follows. Diffusion maps techniques are based on a Markov random walk on the graph of the data.when the random walk for a number of time steps occurs, a measure for the proximity of the datapoints is obtained. By using this measure, the diffusion distance an define. In the low-dimensional representation of the data, the pairwise diffusion distances are kept as well as possible. In the diffusion maps, first the graph of the data is constructed. The weights of the edges in this constructed graph are computed using the Gaussian kernel function, leading to a matrix W with the following entries

$$W_{ij} = e^{\frac{-\|x_i - x_j\|^2}{2\varsigma^2}} \tag{1}$$

where $\varsigma^2$ indicates the variance of the Gaussian. Afterwards, the normalization process of the matrix $W$ is performed in a way that its rows add up to 1. In this way, a matrix $P^{(1)}$ is constructed with entries

$$P_{ij}^{(1)} = \frac{W_{ij}}{\sum_k W_{ik}} \tag{2}$$

Since diffusion maps techniques originate from dynamical systems, the resulting matrix $P^{(1)}$ is regarded a Markov matrix that defines the forward transition probability matrix of a dynamical process. So, the matrix $P^{(1)}$ shows the probability of a transition from one datapoint to another datapoint in one time step. The forward probability matrix for t time steps $P^{(t)}$ can given by $(P^{(1)})^t$. By using the random walk forward probabilities $P_{ij}^{(t)}$, the diffusion distance is computed by

$$D^{(t)}(x_i, x_j) = \sqrt{\sum_k \frac{(P_{ik}^{(t)} - P_{jk}^{(t)})^2}{\varphi(x_k)^0}} \tag{3}$$

In the equation (3), $\varphi(x_k)^0$ is for attributing more weights to parts of the graph with the high density. It is can defined by $\varphi(x_i)^0 = \frac{m_i}{\sum_j m_j}$, where $m_i$ is the degree of node $x_i$ omputed by $m_i = \sum_j P_{ij}$. From Equation (3), it can be concluded that pairs of data points with a high forward transition probability have a small diffusion distance. The main idea behind the diffusion distance is that it is based on many paths in the graph. This makes the diffusion distance more robust to noise than the geodesic distance. In the low-dimensional representation of the data $Y$, diffusion maps attempt to keep the diffusion distances. It can be shown by spectral theory on the random walk that the low-dimensional representation $Y$ that retains the diffusion distances is constructed by the $d$ nontrivial principal eigenvectors of the eigen problem

$$P^{(t)} v = \lambda v \tag{4}$$

Because the graph is fully connected, the largest eigenvalue is minor (viz. $\lambda_1 = 1$), its eigenvector $v_1$ is thus discarded. So The low-dimensional representation $Y$ is given by the next d principal eigenvectors. In the low dimensional representation Y, the eigenvectors can normalized by their corresponding eigenvalues. Therefore, the low-dimensional data representation Y is given by

$$Y = \{\lambda_2 v_2, \lambda_3 v_3, \ldots \lambda_{d+1} v_{d+1}\} \tag{5}$$

### 3.1. Diffusion maps on Synthetic Data

In order to examine the performance of algorithms, we first evaluate algorithms for dimensionality reduction task only. We test on two different standard datasets:

### 3.1.1. Swiss Roll Example

A randomly sampled plane is rolled up into a spiral. A good nonlinear dimensionality reduction technique should unroll this example into a plane. If the number of nearest neighbors is set to large, the embedding will across folds. This dataset was proposed by Tannenbaum [24]. Figure (2) shows this dataset with N=2000 data points.

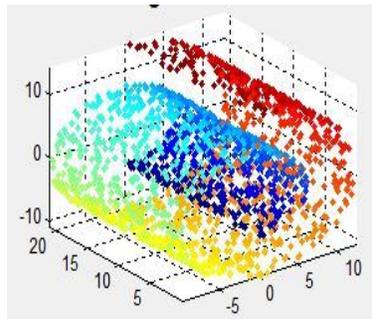

Figure 2. Swiss Roll dataset for N=2000 data points

### 2.1.2. Punctured Sphere Example

The bottom ¾ of a sphere is sampled non-uniformly. The sampling is densest along the top rim and sparsest on the bottom of the sphere. The 2D projection should display concentric circles. The user can control the height of the sphere. This dataset was written by Saul [25]. Figure (3) shows this dataset with N=2000 data points.

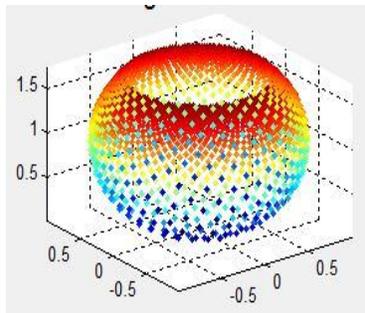

Figure 3. Punctured sphere dataset for N=2000 data points

### 3.1.3. Dimensionality reduction results on two datasets

We tested DM method on two datasets that explained before. Figure (4) and Figure (5) show the embedding results of this method. We set Sigma={1, 4, 10} in both datasets. The performance of DM on Sigma=10 is the best in both datasets.

International Journal in Foundations of Computer Science & Technology (IJFCST), Vol.4, No.6, November 2014

| Diffusion maps | Sigma=1 (a) | Sigma=4 (b) | Sigma=10 (c) |
|---|---|---|---|
| | | | |

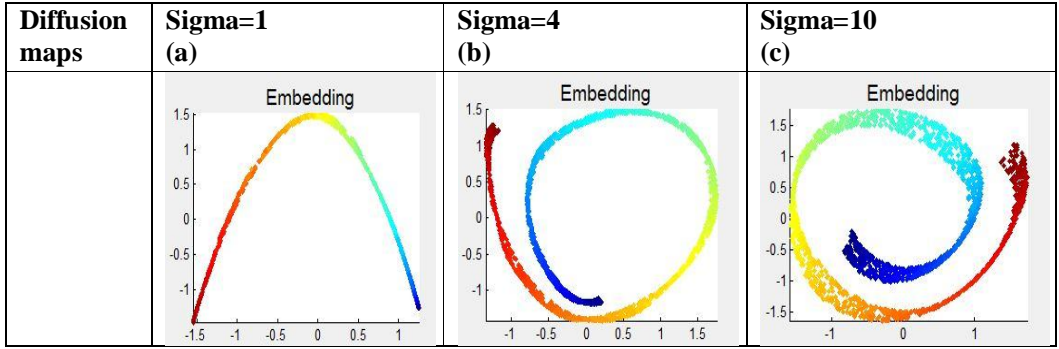

Figure 4. Dimensionality result on Swiss Roll dataset: (a) Sigma=1; (b) Sigma=4; (c) Sigma=10.

| Diffusion maps | Sigma=1 (a) | Sigma=4 (b) | Sigma=10 (c) |
|---|---|---|---|
| | | | |

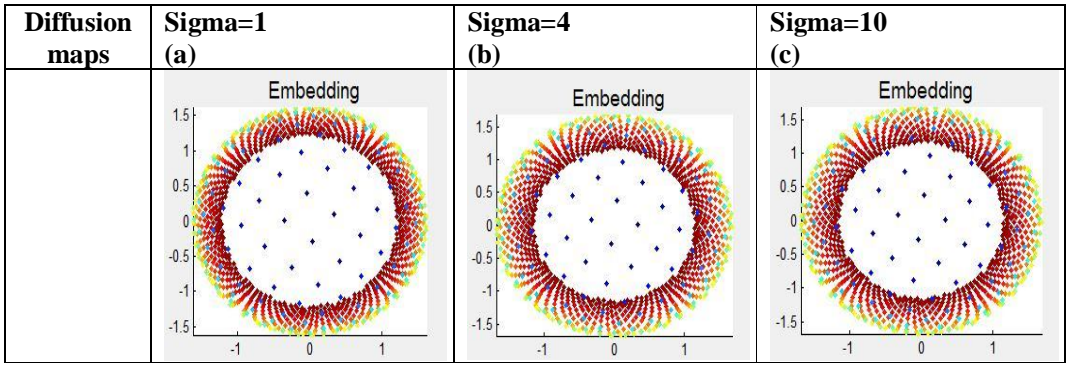

Figure 5. Dimensionality result on Punctured Sphere dataset: (a) Sigma=1; (b) Sigma=4; (c) Sigma=10.

## 4. EXPERIMENT WITH REAL DATA

### 4.1. Feature Vectors

For evaluating of our manifold learning methods, we tested on three features as follows:

#### 4.1.1. 73-D edge direction histogram:

Edge direction histogram encrypts the distribution of the directions of edges. Edge direction histogram has a total of 73 bins, in which the first 72 bins are the count of edges with directions quantized at five degrees interval, and the last bin is the count of number of pixels that do not contribute to an edge. To pay back for different image sizes, they normalized the entries in histogram as follows:

$$H_i = \begin{cases} \dfrac{H(i)}{M_e}, if\ i \in [0,...,71] \\ \dfrac{H(i)}{M}, if\ i = 72 \end{cases} \quad (6)$$

where $H(i)$ is the count of bin i in the edge direction histogram; $M_e$ is the total number of edge points that detected in the sub-block of an image; and $M$ is the total number of pixels in the sub-block. It uses Canny filter to detect edge points and Sobel operator to compute the direction by the gradient of each edge point [23].

International Journal in Foundations of Computer Science & Technology (IJFCST), Vol.4, No.6, November 2014

#### 4.1.2. 144-D color auto-correlogram (HSV):

The color auto-correlogram was proposed to describe the color distributions and the spatial correlation of pairs of colors together. The histogram has three dimensions. The first two dimensions of it are the colors of any pixel pair and the third dimension is their spatial distance. Let $I$ show the entire set of image pixels and $I_{c(i)}$ represent the subset of pixels with color $c(i)$, then the color auto-correlogram is defined as:

$$r_{ij}^{(k)} = \Pr_{P_1 \in I_{c(i)}, P_2 \in I}[P_2 \in I_{c(i)} \| P_1 - P_2 \| = d] \tag{7}$$

where $i, j \in \{1, 2, ..., K\}$, $d \in \{1, 2, ..., L\}$ and $|p_1 - p_2|$ is the distance between $p_1$ and $p_2$ pixels. Color auto-correlogram reduces the dimension from $O(N^2 d)$ to $O(Nd)$ because it only captures the spatial correlation between identical colors. They quantize the HSV color components into 36 bins and set the distance metric to four odd intervals of $d = \{1, 3, 5, 7\}$. So the color auto-correlogram has a dimension of $144 (36 \times 4)$ [23].

#### 4.1.3. 225-D block-wise color moments (LAB):

The first (mean), the second (variance) and the third order (skew- ness) color moments are efficient and effective in representing the color distributions of images. The first three moments are defined as:

$$\mu_i = \frac{1}{N} \sum_{j=1}^{N} f_{ij} \tag{8}$$

$$\sigma_i = (\frac{1}{N} \sum_{j=1}^{N} (f_{ij} - \mu_i)^2)^{\frac{1}{2}} \tag{9}$$

$$s_i = (\frac{1}{N} \sum_{j=1}^{N} (f_{ij} - \mu_i)^3)^{\frac{1}{3}} \tag{10}$$

where $f_{ij}$ is the value of the $i$'th color component of the image pixel $j$, and $N$ is the total number of pixels in the image.

Color moments offer a very dence representation of image content as compared to other color features. Using of three color moments as described above, only nine components (three color moments, each with three color components) will be used. Due to this density, it may not have good discrimination power. for NUS_WIDE dataset, they extract the block-wise color moments over 5×5 fixed grid partitions, giving rise to a block-wise color moments with a dimension of 225 [23].

### 4.2. Dataset

We performed experiment on NUS-WIDE-LITE dataset that represented by Tat-Seng Chua [23]. This is a large web image dataset downloaded from Flickr. All dataset samples are supervised and labeled with 81 concepts and it has about 2 labels per image. NUS-WIDE-LITE dataset includes a set of 55615 images and their associated tag randomly selected from the full NUS-WIDE dataset. Its cover the full 81 concepts from NUS-WIDE. We prune images with less than five labels and then we work on subdataset. We randomly divide the images into training and testing subset. Half of them are used as training data and the rest are used for testing data. We represent each image as three feature vectors.

### 4.3. Annotation Method and Evaluation Protocol

Since our interest was in the performance of image annotation, we simply use K nearest neighbour method. The system output the most frequent labels in the k retrieved neighbours. For NUS-WIDE-LITE dataset, we tested k={4, 8,16, 32} nearest neighbour and d={10, 20, 30, 40, 50} dimensions for dimensionality reduction methods. For evaluating of annotation, we follow the methodology of previous work that Hideki Nakayama [26] proposed it. The system annotates test image with 5 words for each. These words are then compared with the ground truth.

To facilitate a quantitative comparison, we use publicly available feature files in the experiment. We test on three features: 1) 225-D block-wise color moments, 2) 144-D color auto-correlogram, 3) 73-D edge direction histogram. These features are provided by the authors of [23]. By using the manifold learning algorithms, we decrease the dimensions of these three feature vectors to d={10, 20, 30, 40, 50} dimensions.

### 4.4. Experimental Results

We report the results for the NUS-WIDE-LITE dataset. In this dataset, the dimension of visual features is too large for k nearest neighbourhood method. So we decrease the dimensions of these three feature vectors to d={10, 20, 30, 40, 50} dimensions. First the dimensionality reduction time for each of four methods is summarized in Table (1). Target dimensionality is set to d=30, and we use a 5-core Intel 2.30 GHz system for computation. We compare the performance of DM method with three other dimensionality reduction algorithms; PCA, LLE and LEM. The dimensionality reduction time of PCA is the lowest between all methods and the dimensionality reduction time of DM is the highest between all methods.

Next we summarized the results for image annotation task. We give a comparison of the annotation accuracy (Average Precision) with k=8 in KNN method. Average Precision is originally used in information retrieval systems to evaluate the document ranking performance for query retrieval [27]. In our comparison, DM exhibit superior performance, and achieve comparable and better performance than the other three algorithms.

Because we tested on three features, we show our experimental result of each feature separately. The first feature is 73-D edge direction histogram. The results for annotation with this feature are summarized in Figure (6). On 73-D edge direction histogram, the performance of PCA of is the lowest between all methods. The DM achieves the best performance. The results of LLE and LEM are moderate. The second feature is 144-D color auto-correlogram (HSV). The results for annotation with 144-D color auto-correlogram (HSV) feature are summarized in Figure (7). The third feature is 225-D block-wise color moments (LAB). The DM achieves the best performance. The results for annotation with 225-D block-wise color moments (LAB) feature are summarized in Figure (8). In our comparison on three features, 225-D block-wise color moments (LAB) exhibit superior performance on different dimensions.

Table 1 Dimensionality reduction computation time for each of four methods.

| Run Time (s) | Edge histogram | Color auto-correlogram | Color moments |
|---|---|---|---|
| PCA | 3.36 | 2.12 | 4.48 |
| LLE | 17.79 | 16.04 | 23.97 |
| LEM | 8.89 | 9.06 | 15.85 |
| **DM** | 169.70 | 620.40 | 779.47 |

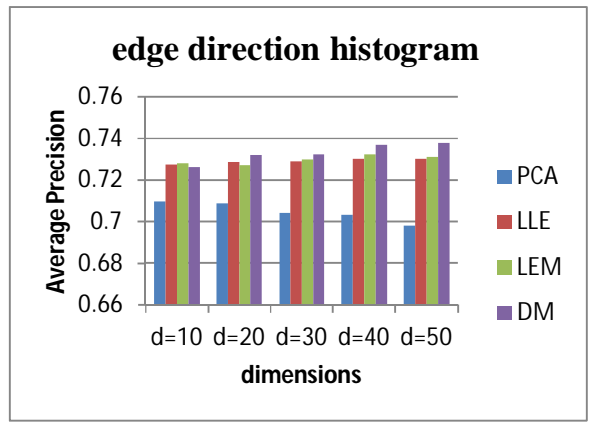

Figure 6. Average Precision for the NUS-WIDE-LITE dataset with different dimensions on edge direction histogram.

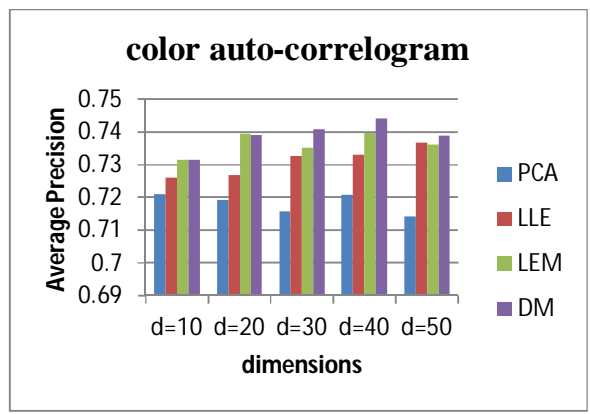

Figure 7. Average Precision for the NUS-WIDE-LITE dataset with different dimensions on color auto-correlogram.

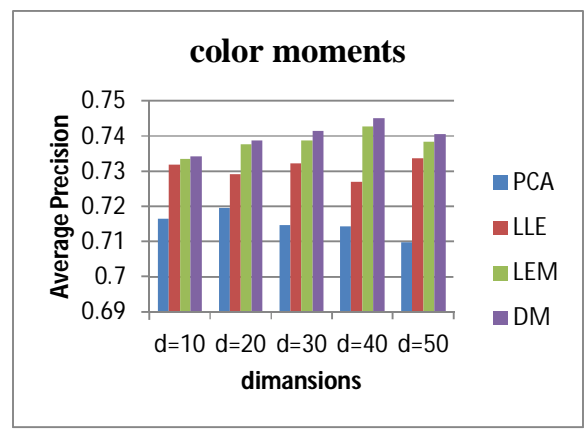

Figure 8. Average Precision for the NUS-WIDE-LITE dataset with different dimensions on color moments.

## 5. CONCLUSION

In this article, we investigated and compared a manifold learning dimensionality reduction algorithm (DM) for image annotation task. Obtaining powerful small algorithms in a scalable manner is an important issue in implementing large-scale image annotation systems. Linear methods like PCA enable training in linear time and are suitable for this purpose. Overall, we can balance the trade-off between computation efficiency and annotation accuracy by selecting the dimensions of feature vectors. We can obtain a small-dimension latent subspace reflecting the semantic distance of instances. We used DM method for this goal and we achieve the best performance in value of Average precision. Our future work could be extended to other compression formats and the other advanced manifold learning algorithms.

## REFERENCES

[1] E.Y.Chang, K.Goh, G.Sychay, and G.Wu, (2003) "CBSA: content-based soft annotation multimodal image retrieval using Bayes point machines" , IEEE Trans. Circuits and Systems for Video Technology, 13(1): pp26-38.

[2] C.Cusano, G. Ciocca, and R.Schettini, (2004) "Image annotation using SVM", In proc. SPIE Conference on Internet Imaging IV, volume SPIE.

[3] P.Duygulu, K.Barnard, N. de Freitas, and D. Forsyth, (2002), "Object recognition as machine translation: Learning a lexicon for a fixed image vocabulary", In Proc. ECCV, pp97-122.

[4] S.Feng, R. Manmatha, and V. Lavrenko, (2004) "Multiple Bernoulli relevance models for image and video annotation", In proc. IEEE CVPR, volume 2, pp1002-1009.

[5] V.Larvenko, R. Manmatha, and J. Jeon, (2003) "A model for learning the semantics of pictures", In proc. NIPS.

[6] J.Liu, M. Li, Q. Liu, H. Lu, and S. Ma, (2009) "Image annotation via graph learning", Pattern Recognition,42(2), pp218-228.

[7] J.Liu, M. Li, W.-Y. Ma, Q. Liu, and H. Lu, (2006) "An adaptive graph model for automatic image annotation", In proc. ACM MIR, pp61-70.

[8] D.M. Blei and M. I. Jordan, (2003) "Modeling annotated data", In Proc. ACM SIGIR, pp127-134.

[9] F.Monay and D. Gatica-Perez, (2004) "PLSA-based image auto-annotation: constraining the latent space", In Proc. ACM multimedia.

[10] G.Carneiro, A. B. Chan, P. J. Moreno, and N. Vasconcelos, (2007) "Supervised learning of semantic classes for image annotation and retrieval", IEEE Trans. Pattern Analysis and Machine Intelligence, 29.

[11] A.Makadia, V. Pavlovic, S.Kumar, (2008) "A new baseline for image annotation", In proc. ECCV, pp316-329.

[12] M.Guillaumin, T. Mensink, J. Verbeek, and C. Schmid, (2009) "Tagprop: Discriminative metric learning in nearest neighbor models for image auto-annotation", In Proc. IEEE ICCV, pp309-316.

[13] Xiaoming Huo, Xuelei Ni, and Andrew K.Smith, (2004) "A Survey of Manifold based Learning Methods", INFORMS Annual Meeting, CO.USA.

[14] Nathan Mekuz, Christian Bauckhage and John K. Tsotsos, (2009) "Subspace manifold learning with sample weights", Image and Vision Computing 27, pp.80–86.

[15] B. Raducanu and F.Dornaika, (2012) "A supervised non-linear dimensionality reduction approach for manifold learning", Pattern Recognition 45, pp2432–2444

[16] Carlotta Orsenigo and Carlo Vercellis, (2013) "A comparative study of nonlinear manifold learning methods for cancer microarray data classification", Expert Systems with Applications 40, pp2189–2197.

[17] Abdenour Hadid and MattiPietik ainen, (2013) "Demographic classification from face videos using manifold learning", Neurocomputing 100 , pp197–205.

[18] Yahong Han, ZhongwenXu, ZhigangMa and ZiHuang, (2013) "Image classification with manifold learning for out-of- sample data", Signal Processing 93 , pp2169–2177

[19] Xianming Liu, Rongrong Ji, Hongxun Yao, Pengfei Xu, Xiaoshuai Sun and Tianqiang Liu, (2008) "Cross-media manifold learning for image retrieval and annotation" MIR, '08, Proceedings of the 1st ACM international conference on Multimedia information retrieval, pp141-148.

[20] L.J.P. van der Maaten, E.O. Postma and H.J. van den Herik, (2007) "Dimensionality Reduction: A Comparative Review", Journal of Machine Learning Research, JMLR.

[21] M.H. Law and A.K. Jain, (2006) "Incremental nonlinear dimensionality reduction by manifold learning", IEEE Transactions of Pattern Analysis and Machine Intelligence, 28(3) pp377–391.

[22] A.Ng, M. Jordan, and Y. Weiss, (2001) "On spectral clustering: Analysis and an algorithm", In Advances in Neural Information Processing Systems, Cambridge, MA, USA , , volume 14, pp 849–856

[23] T.-S. Chua, J. Tang, R. Hong, H. Li, Z. Luo, and Y. -T. Zheng, (2009) "NUS-WIDE: a real-word web image database from National University of Sangapore", In proc. ACM CIVR.

[24] Joshua B. Tenenbaum, Vin de Silva and John C. Langford, (2000) "A Global Geometric Framework for Nonlinear Dimensionality Reduction", Science 290.

[25] Lawrence K. Saul and Sam T. Roweis, (2003) "Think Globally, Fit Locally: Unsupervised Learning of Low Dimensional Manifolds", Journal of Machine Learning Research 4, pp119-155.

[26] H.Nakayama, T. Harada, and Y. Kuniyoshi, (2010) "Evaluation of dimensionality reduction methods for image auto-annotation", British Machine Vision Association,  pp1-12.

[27] Min-Ling Zhang and Zhi-Hue Zhou, (2007) "ML-KNN: A lazy learning approach to multi-label learning", Pattern Recognition 40, pp2038-2048.